\documentclass[conference]{IEEEtran}
\IEEEoverridecommandlockouts
\usepackage{cite}
\usepackage{amsmath,amssymb,amsfonts}
\usepackage{algorithmic}
\usepackage{graphicx}
\usepackage{textcomp}
\usepackage{xcolor}
\def\BibTeX{{\rm B\kern-.05em{\sc i\kern-.025em b}\kern-.08em
    T\kern-.1667em\lower.7ex\hbox{E}\kern-.125emX}}
\begin{document}

\title{Heuristics and Biases in AI Decision-Making: Implications for Responsible AGI}

\author{\IEEEauthorblockN{1\textsuperscript{st} Payam Saeedi}
\IEEEauthorblockA{\textit{Golisano Institute for Sustainability} \\
\textit{Rochester Institute of Technology}\\
Rochester, The United States \\
ps4019@rit.edu}
\and
\IEEEauthorblockN{2\textsuperscript{nd} Mahsa Goodarzi}
\IEEEauthorblockA{\textit{Information Sciences \& Technology, CEHC} \\
\textit{University at Albany, SUNY}\\
Albany, The United States \\
mgoodarzi@albany.edu}
\and
\IEEEauthorblockN{3\textsuperscript{rd} M Abdullah Canbaz}
\IEEEauthorblockA{\textit{Information Sciences \& Technology, CEHC} \\
\textit{University at Albany, SUNY}\\
Albany, The United States \\
mcanbaz@albany.edu}

 }

\maketitle

\begin{abstract}
We investigate the presence of cognitive biases in three large language models (LLMs): GPT-4o, Gemma 2, and Llama 3.1. The study uses 1,500 experiments across nine established cognitive biases to evaluate the responses and consistency of the models. GPT-4o demonstrated the strongest overall performance. Gemma 2 showed strengths in addressing the sunk cost fallacy and prospect theory; however, its performance varied across different biases. Llama 3.1 consistently underperformed, relying on heuristics and exhibiting frequent inconsistencies and contradictions. The findings highlight the challenges of achieving robust and generalizable reasoning in LLMs, and underscore the need for further development to mitigate biases in artificial general intelligence (AGI). The study emphasizes the importance of integrating statistical reasoning and ethical considerations in future AI development.
\end{abstract}


\section{Introduction}\label{intro}

Cognitive biases and heuristics are well-established phenomena of the human mind, shaping how individuals process information, make judgments, and make decisions. These biases emerge from heuristics — mental shortcuts that simplify complex tasks by substituting them with cognitively easier alternatives \cite{todd_environments_2007}. While heuristics enable quick and efficient reasoning, they also introduce systematic errors that impact judgment and decision-making \cite{ehrlinger_decision-making_2016,murata_influence_2015,tversky_judgment_1974}. Understanding whether such biases, embedded in the data and interactions that shape Large Language Models (LLMs), are reflected in their outputs is not only critical for evaluating their alignment with human cognition but also vital for the development of Artificial General Intelligence (AGI). AGI, envisioned as systems capable of performing any intellectual task a human can, must navigate the intricacies of human-like reasoning while avoiding harmful or irresponsible biases. Measuring, analyzing, and responsibly implementing these biases in LLMs can guide the design of AGI that mirrors the nuanced nature of human cognition without replicating its systemic errors, ensuring both ethical integrity and practical utility.

Research on cognitive biases in AI systems is still in its infancy. However, prior studies have explored how human cognitive biases, such as those influencing judgment and decision-making, might manifest in AI and machine learning systems due to similarities in their training processes \cite{g_harris_mitigating_2020,lin_mind_2023,martinez-human2022}. Scholars have called for interdisciplinary approaches to analyze these biases, emphasizing their relevance for explainable AI (XAI) and algorithmic behavior \cite{bertrand_how_2022,irving_ai_2019,martinez-human2022}. Notably, humans have been shown to outperform traditional machine learning algorithms in addressing certain biases, such as loss aversion and overconfidence \cite{blohm_its_2022}, underscoring the complexity of replicating nuanced human reasoning in computational systems.

Cognitive biases are also present within LLMs and natural language generation tasks, with evidence observed in areas such as code generation \cite{jones_capturing_2022}, sentiment analysis \cite{abramski_cognitive_2023}, cognitive reflection tests (CRT) \cite{hagendorff_why_2023}, and decision-making \cite{ma_is_2023}. Despite being designed to simulate rationality through mathematical and computational principles, LLMs often exhibit human-like responses to biases, including framing effects, anchoring, availability heuristics, and attribute substitution. For instance, experiments with models such as Codex, CodeGen, and GPT-3 have revealed errors related to these biases during code generation \cite{jones_capturing_2022}. Similarly, ChatGPT demonstrated a high tendency to emulate human-like reasoning when exposed to biases such as the gambler’s fallacy, zero-cost effect, ultimatum game, endowment effect, and framing effect \cite{azaria_chatGPT_2023,ma_is_2023}. Nevertheless, newer iterations of GPT, such as GPT-3.5 and GPT-4, have shown improvements in addressing cognitive biases, particularly in chain-of-thought (system 2) prompting compared to zero-shot (system 1) approaches \cite{hagendorff_human-like_2023}.

Despite these advancements, inconsistencies in model responses persist, especially when identical or similar prompts are repeated. Such inconsistencies, often described as “hallucinations,” reflect the generation of incorrect or contradictory outputs \cite{10.1007/978-3-031-53468-3_17}. However, researchers have debated the appropriateness of this term \cite{hicks_chatgpt_2024}, as these inconsistencies occasionally yield accurate responses. Including repetition in performance evaluations enhances robustness and reproducibility \cite{zhu_evaluation_2024}, yet most existing studies fail to quantify the inconsistency of responses generated across multiple iterations of a prompt \cite{abramski_cognitive_2023,irving_ai_2019,gal_cognitive_2023,macmillan-scott_irrationality_2024}.

Understanding how biases inherent in human cognition might have transferred to LLMs through training data and interactions is a critical question \cite{kudless_hierarchies_2023}. Additionally, it is essential to distinguish between the nature of biases in humans and AI systems \cite{gulati_biased_2023}. Some researchers have advocated for treating LLMs as active participants in cognitive experiments to explore these biases, mirroring methodologies used in human cognition studies \cite{sartori_language_2023}.

In this paper, we adopt a rigorous experimental framework to evaluate cognitive biases in production-grade LLMs. Specifically, we expose three models, namely GPT-4o, Gemma 2 and Llama 3.1, to a diverse set of well-established cognitive biases through carefully designed word experiments, systematically measuring their responses across multiple iterations of identical and varied prompts. Our study seeks to address the following research questions:
\begin{itemize}
    \item  To what extent do cognitive biases inherent in human reasoning, such as loss aversion, framing effects, and the conjunction fallacy, manifest in the responses of GPT-4o, Gemma 2, and Llama 3.1 in probabilistic and decision-making scenarios?
    \item How consistent are the responses of GPT-4o, Gemma 2, and Llama 3.1 when exposed to identical prompts targeting the same cognitive biases, and what factors contribute to observed variations?
    \item Can GPT-4o, Gemma 2, and Llama 3.1 recognize the presence of cognitive biases in their decision-making processes, and how effectively do they apply statistical reasoning to mitigate these biases?
\end{itemize}

By addressing these questions, we aim to advance the understanding of how LLMs process cognitive tasks and contribute to the broader objective of developing responsible Artificial General Intelligence (AGI). The findings of this study are crucial for identifying the limitations and strengths of LLMs in reasoning under uncertainty and for guiding their development as ethical and effective decision-making tools.

\section{Experiment Settings: Investigating Cognitive Biases in LLMs}

Machine learning and deep learning models operate fundamentally differently from the human mind, which limits the direct applicability of certain cognitive heuristics. For instance, asking ChatGPT to "recall" six or twelve instances of its own assertive behavior during interactions with colleagues \cite{schwarz_ease_nodate} would be infeasible, as these models lack personal experience or episodic memory. To explore the presence of cognitive biases in Large Language Models (LLMs), we systematically reviewed an extensive list of biases and selected nine that could be operationalized as prompts designed to elicit machine-readable responses. 

Below, we describe each selected bias, its theoretical foundations, and its relevance to understanding the reasoning patterns of LLMs.

\textbf{Loss Aversion:} Loss aversion is the tendency to perceive losses as more impactful and painful than equivalent gains are pleasurable. This bias leads individuals to avoid losses with greater motivation than pursuing gains of similar value \cite{kahneman_prospect_1977, yechiam_acceptable_2019}. In our experiments, we designed prompts requiring the three models to navigate scenarios involving trade-offs between potential losses and gains, analyzing whether their responses prioritized loss avoidance.

\textbf{Sunk Cost Fallacy:} The sunk cost fallacy describes an irrational commitment to a course of action based on prior investments of time, money, or effort, even when abandoning the action would yield greater benefits. The stronger the perceived investment, the harder it becomes to disengage \cite{arkes_psychology_1985, arkes_sunk_1999}. We tested whether the models exhibited this bias by presenting scenarios that juxtaposed past investments with future payoffs.

\textbf{Prospect Theory / Risk Aversion:} Prospect theory explains risk aversion as the tendency to prefer certain outcomes over uncertain ones, even when the uncertain option offers a potentially higher reward. This behavior is often driven by an aversion to ambiguity and a desire for safety \cite{kahneman_prospect_1977, kahneman_choices_1984, mcdermott_evolutionary_2008}. Prompts targeting this bias evaluated whether the three studied models adhered to probabilistic reasoning or mirrored human tendencies to avoid uncertainty.

\textbf{Framing Effect:} The framing effect occurs when decisions are influenced by whether information is presented positively (gain) or negatively (loss), despite identical expected outcomes \cite{tversky_framing_1981, Tversky1985, thomas_reducing_2012}. By crafting scenarios with alternative framings, we examined whether our three models produced consistent responses or displayed susceptibility to this bias. In this case, we asked each model whether they would support or not support an experimental life-saving medicine that would:

\begin{enumerate}
    \item Save 300 out of 800 patients
    \item Kill 500 out of 800 patients
\end{enumerate}

\textbf{Halo Effect:} The halo effect describes the tendency to form an overly favorable impression of one attribute of a person or entity based on unrelated positive traits \cite{schouteten_influence_2019, nisbett_halo_1977}. This bias often results in skewed evaluations. Prompts were designed to test whether the models demonstrated similar overgeneralization in their reasoning.

\textbf{Conjunction Fallacy (The Linda Problem):} The conjunction fallacy is the erroneous belief that the likelihood of two events occurring together is higher than the probability of either event occurring alone. This violates fundamental probability principles \cite{moro_nature_2009, tversky_extension_1983}. We presented the three models with variations of the "Linda Problem" to assess whether they could avoid this bias by applying statistical reasoning.

\textbf{Bias of Resemblance / Representativeness:} This bias involves estimating the likelihood of an event based on its resemblance to a prototype or stereotype, often neglecting statistical base rates \cite{tversky_judgment_1974, moro_nature_2009}. We crafted prompts to determine whether the models defaulted to representativeness heuristics or leveraged statistical reasoning.

\textbf{Neglect of Probability 1 \& 2: }
Neglect of probability occurs when the likelihood of an event is ignored or downplayed due to emotional or vivid factors. This bias manifests in two forms: (1) comparing scenarios based on perceived magnitude rather than probability, and (2) treating independent events as correlated \cite{schade_probability_2004, kahneman_thinking_2013}. Experiments explored whether the examined models consistently adhered to probabilistic reasoning across these scenarios.

\subsection*{Experimental Design and Procedure}

\textbf{Data:} 
We examined three models released in 2024, namely, GPT-4o \cite{noauthor_openai_nodate}, Llama 3.1 \cite{noauthor_meta-llamallama-31-405b_nodate}, and Gemma 2 \cite{noauthor_gemma_nodate}, and provided each with ten identical prompts for each cognitive bias. 

The prompts were adapted from psychological experiments which used similar text to identify cognitive biases in the first place.

To ensure consistency, zero-shot chain-of-thought prompting was used, and the models were assigned the role of "a human participating in a social experiment" \cite{sartori_language_2023}. Settings available to the general public through chat-bots were preserved, with the temperature kept at its default value.

At the conclusion of each prompt, the model was instructed to explain its reasoning and select one of two alternatives. Except for the framing effect wherein the model is asked if they support or not support a scenario, responses were labeled as follows:
\begin{itemize}
    \item \textbf{Elaborate:} Responses based on detailed statistical reasoning.
    \item \textbf{Intuitive:} Responses reflecting simplifying heuristics.
    \item \textbf{Other:} Non-responses or deviations from the two answers offered above. 
\end{itemize}

For example, a prompt described an individual who was tidy, organized, and detail-oriented, while also being a survivor of domestic violence. The model was then asked whether this individual was more likely to be a librarian or both a librarian and an advocate for domestic violence survivors. Statistically, the probability of belonging to both sets (librarian and advocate) is lower than the probability of belonging to either set individually. However, reliance on heuristic reasoning often leads to the intuitive yet incorrect conclusion that the individual is both a librarian and an advocate. Identifying the individual as solely a librarian was deemed the elaborate response.

Inspired by the experimental designs used to study these biases in humans, we replicated each experiment 150 times and tallied the proportion of elaborate and intuitive responses for each prompt. Responses were evaluated against a curated list of keywords associated with each bias. Keywords were extracted from prominent literature \cite{mcdermott_evolutionary_2008, moro_nature_2009, schade_probability_2004, sun_evolution_2020, tversky_judgment_1974, yechiam_acceptable_2019} to determine whether the models recognized or referenced the biases in their reasoning.

For reproducibility purposes, we have made the data curated and the model parameters along with prompts and sample results available for the community in our GitHub page\cite{noauthor_github_nodate}.

\section{Experimental Results}
We organized our experimental results to address the three research questions outlined in Section \ref{intro}. Each subsection corresponds to one of the research questions, providing a detailed analysis of how the models responded to cognitive biases, their consistency and variability, and their ability to recognize and mitigate these biases through statistical reasoning. The results are structured to facilitate a comparative understanding of GPT-4o, Gemma 2, and Llama 3.1 across the nine cognitive biases examined in this study.

\subsection{Bias Manifestation}
\begin{figure}[!h]
  \includegraphics[width=\columnwidth]{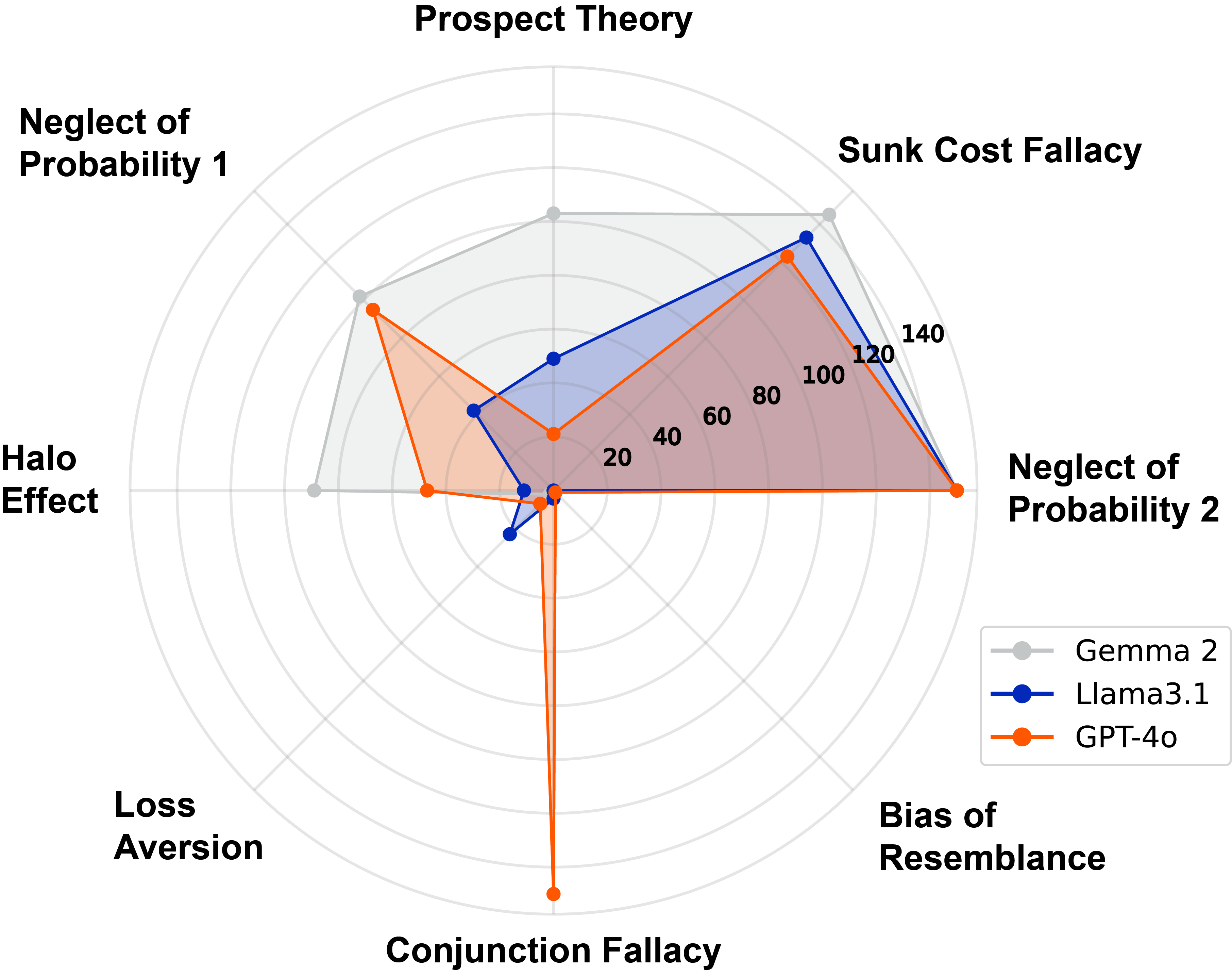}
  \caption{Elaborate responses provided by GPT-4o, Gemma 2, and Llama 3.1.}
  \label{fig:spider chart}
\end{figure}

To evaluate the extent to which cognitive biases manifest in language model responses, 1,500 experiments were conducted across nine cognitive biases using identical prompts for GPT-4o, Gemma 2, and Llama 3.1. The radar plot in Figure~\ref{fig:spider chart} provides a comparative view of the elaborate responses generated by the models in question, across various cognitive biases. Each axis represents one cognitive bias, and the distance from the center reflects the frequency of elaborate (statistically reasoned) responses. The areas enclosed by the lines for each model highlight their relative strengths and weaknesses in addressing these biases. Below, we discuss the key insights observed from the results.

\textbf{Conjunction Fallacy - A Dominant Strength for GPT-4o:} 
GPT-4o exhibits an exceptionally high performance in recognizing and reasoning through the conjunction fallacy, with its response rate far surpassing those of Gemma 2 and Llama 3.1. Its ability to correctly identify and explain why the probability of two events occurring together is always less than or equal to the probability of either event alone underscores its strength in applying fundamental statistical principles. 

In contrast, both Gemma 2 and Llama 3.1 display significant limitations in this domain, reflecting a reliance on heuristic-based reasoning that frequently leads to incorrect responses. This disparity highlights GPT-4o's superior capacity for handling probabilistic reasoning tasks, making it a potentially reliable tool for applications where logical rigor in decision-making is paramount. However, its outlier performance in this bias also raises questions about why similar statistical principles do not generalize as effectively across other biases.

\textbf{Sunk Cost Fallacy and Prospect Theory - Strengths of Gemma 2:} 
Gemma 2 demonstrates notable proficiency in addressing the sunk cost fallacy and prospect theory, outperforming GPT-4o and Llama 3.1. Its relatively higher scores suggest a better alignment with human-like reasoning when evaluating decisions influenced by past investments (sunk cost) and risk preferences (prospect theory). This performance may stem from Gemma 2's design, which appears to emphasize contextual sensitivity in scenarios requiring nuanced judgments about losses and risks.

Despite these strengths, Gemma 2’s overall performance across other biases remains inconsistent, indicating that its capabilities may be domain-specific rather than generalizable across a broad spectrum of cognitive tasks.

\textbf{Neglect of Probability - Persistent Challenges Across Models:} While all three models correctly acknowledged the independence of probabilistic events targeted in neglect of probability 2; the neglect of probability 1 bias presents a persistent challenge for all of them. None were achieving consistently high levels of elaborate responses in the latter case. GPT-4o and Gemma 2 perform moderately, while Llama 3.1 struggles significantly. This finding suggests that even advanced models frequently fail to incorporate probabilistic reasoning when the context involves emotional or vivid imagery, a hallmark of this bias. 

The inability to consistently recognize or mitigate probability neglect indicates an inherent limitation in the models' design, particularly in bridging the gap between statistical principles and emotionally charged or vividly framed scenarios. Addressing these deficiencies is crucial for improving the reliability of AI in high-stakes decision-making.

\textbf{Bias of Resemblance - A Common Weakness:} All three models exhibit significant struggles with the bias of resemblance, with GPT-4o and Gemma 2 defaulting to stereotypical reasoning in a majority of cases. This bias demonstrates the models' propensity to rely on surface-level similarities rather than statistical base rates, a behavior that mirrors human heuristic-driven reasoning. Llama 3.1, in particular, shows minimal ability to overcome this bias, further emphasizing its reliance on simplistic heuristics.

The widespread difficulty across models in addressing this bias underscores the need for enhanced mechanisms to prioritize statistical reasoning over superficial associations, especially in applications requiring equitable and unbiased decision-making.

\textbf{Framing Effects - Susceptibility to Presentation Context:} 
The framing effect reveals moderate to high susceptibility among all three models, with GPT-4o, Gemma 2, and Llama 3.1 providing contradictory responses to positive and negative framings of identical scenarios in 57\%, 74\%, and 85\% of cases, respectively. This variability suggests that the models are influenced by the presentation context, failing to maintain consistent reasoning across different framings of the same problem.

Such susceptibility poses challenges for applications where consistency and impartiality in decision-making are critical. While GPT-4o demonstrates relatively lower susceptibility compared to Gemma 2 and Llama 3.1, its moderate performance indicates room for improvement in ensuring contextual robustness.

\textbf{Llama 3.1 - The Least Reliable Performer:} 
Across all biases, Llama 3.1 consistently underperforms, with low rates of elaborate responses and frequent reliance on heuristics. Its inability to provide consistent or statistically sound reasoning highlights fundamental limitations in its architecture or training. Additionally, Llama 3.1’s frequent non-responses or contradictions further diminish its reliability, rendering it unsuitable for tasks requiring high levels of cognitive complexity or probabilistic reasoning.

\textbf{Broader Implications:} 
The comparative analysis presented in the radar plot offers a comprehensive view of the models' reasoning capabilities. GPT-4o emerges as the strongest performer overall, particularly in conjunction fallacy and other tasks requiring probabilistic reasoning. However, its weaknesses in biases such as framing effects and neglect of probability highlight the need for further refinement. Gemma 2, while showing strengths in specific domains such as sunk cost fallacy and prospect theory, struggles with generalizability across other biases. Llama 3.1’s consistently poor performance emphasizes the limitations of simplistic heuristic-driven models in handling complex cognitive tasks.

\subsection{Analysis of Consistency in Responses Across Cognitive Biases}

\begin{figure*}[!h]
  \includegraphics[width=.94\textwidth]{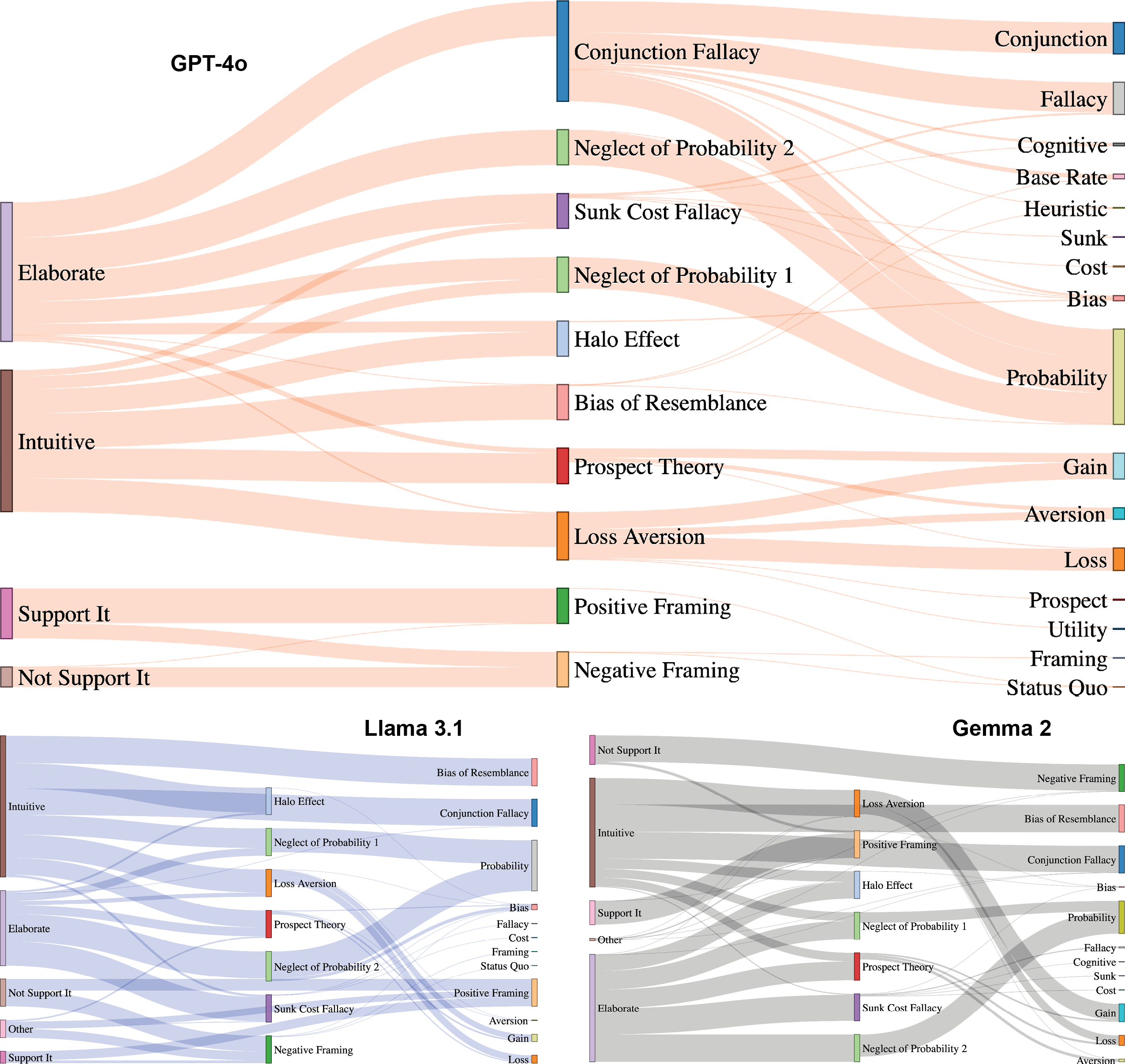}
  \centering
  \vspace{-4mm}
  \caption{Sankey Diagram for bias categories based on responses and bias keywords. Center: GPT-4o, Bottom Left: Llama 3.1, Bottom Right: Gemma 2}
  \vspace{-4mm}
  \label{fig:sankey}
\end{figure*}

The consistency of responses across identical prompts was analyzed to assess the robustness and reasoning reliability of each model. Figure~\ref{fig:sankey} provides a detailed visualization of the flow between response types (\textit{elaborate} and \textit{intuitive}), bias categories, and related keywords in the reasoning chain. These Sankey diagrams highlight how the reasoning paths differ across models and biases, shedding light on key strengths, weaknesses, and inconsistencies.

\textbf{Response Patterns and Variability:} GPT-4o exhibited a relatively well-defined flow in its reasoning, with a notable preference for elaborate responses in biases such as conjunction fallacy, sunk cost fallacy, and Neglect of Probability 2. However, significant variability emerged in biases involving halo effect and neglect of probability 1, where it frequently alternated between elaborate and intuitive reasoning when exposed to identical prompts. This inconsistency suggests that GPT-4o's decision-making in these cases is context-sensitive, heavily influenced by prompt phrasing or subtle variations in the input structure. A similar case applies to inconsistent support for negative framing.

Gemma 2 demonstrated moderate variability, characterized by a tendency to decline responses in scenarios with ambiguous framing or incomplete information. In such cases, Gemma 2 occasionally produced outputs labeled as \textit{indecisive} or irrelevant, reflecting a hesitancy to commit to reasoning paths. Despite this inconsistency, Gemma 2 displayed structured flow for certain biases, such as neglect of probability 2, where its reliance on contextual cues often led to relatively consistent reasoning. However, occasional deviations involving the use of emojis or humor further emphasize its lack of robustness in scenarios requiring strict logical rigor.

Llama 3.1 exhibited the highest level of inconsistency and unpredictability, with frequent transitions between response types. As illustrated in the associated Sankey diagram, Llama 3.1's reasoning paths often resulted in non-responses or contradictions. For instance, in several cases, it explicitly stated one answer (e.g., \textit{“The answer is A”}) but provided reasoning that supported the opposite conclusion (e.g., \textit{“B”}). Additionally, Llama 3.1 struggled to generate coherent flows for biases such as neglect of probability 1, prospect theory, and framing effects, where it failed to maintain logical consistency across prompts.

\textbf{Insights into Bias-Related Reasoning:} The Sankey diagrams reveal distinct patterns in how the models process biases. For instance:

\textit{Conjunction Fallacy:} The strongest flow of elaborate responses was observed for this bias in GPT-4o, which consistently identified and addressed the probabilistic principles involved. This reflects a higher capability for structured and rule-based reasoning. In contrast, Gemma 2 and Llama 3.1 had the strongest flow of intuitive responses for this bias.

\textit{Sunk Cost Fallacy:} Both GPT-4o and Gemma 2 displayed a relatively strong flow of elaborate responses, indicating a capacity to recognize the irrationality of pursuing sunk investments. However, inconsistencies emerged when models failed to consider nuanced contextual factors.

\textit{Neglect of Probability 1:} All three models exhibited inconsistent fragmented flows in this category, with frequent intuitive responses linked to emotionally vivid prompts. This underscores the difficulty AI models face in bridging the gap between statistical reasoning and emotionally charged scenarios.

\textit{Framing Effect:} The reasoning paths for framing effects reveal substantial variability, with frequent contradictions. While Gemma 2 almost always advocates against the negative framing of a scenario, Llama 3.1 refused to support the same context painted in a positive light in over 44\% of the framing experiments. GPT-4o on the other hand, was comfortable in its support for an experimental life-saving drug which would diminish 500 member of a population in 43\% of instances. The lack of a dominant flow to elaborate responses indicates that the presentation context heavily influences reasoning consistency.

\textbf{Keyword Analysis and Reasoning Flow:} 
The Sankey diagrams in figure \ref{fig:sankey} also highlights the presence of bias-related keywords in the reasoning chains, providing insight into how models process specific prompts:
\begin{itemize}
    \item GPT-4o frequently referenced probabilistic terms such as \textit{“expected value”} and \textit{“probability,”} particularly for biases like conjunction fallacy, sunk cost fallacy, and neglect of probability. However, it struggled to integrate these concepts in scenarios involving the bias of resemblance and framing effects, where keywords associated with heuristics often dominated its responses.
    \item Gemma 2 demonstrated reliance on contextual and emotional terms for biases such as loss aversion and prospect theory. While this often led to human-like reasoning, it also contributed to its variability, particularly when prompts required strict adherence to statistical principles. This led to occasional intuitive reasoning in the case of neglect of probability 1.
    \item Llama 3.1 showed the weakest association with bias-related keywords, often failing to reference critical concepts in its reasoning. Its fragmented flows highlight a reliance on surface-level heuristics rather than deeper logical processing.
\end{itemize}

\textbf{Implications for Model Robustness}
The variability and inconsistencies observed across models underscore key challenges in achieving robustness in AI reasoning. GPT-4o\'s relatively strong performance in conjunction fallacy and other rule-based tasks highlights its potential for applications requiring probabilistic reasoning. However, its variability in biases such as framing effects and neglect of probability raises concerns about its adaptability to diverse scenarios.

Gemma 2\'s contextual sensitivity, while advantageous in certain domains, often leads to unpredictable reasoning paths, limiting its applicability in scenarios requiring consistent logical rigor. Llama 3.1's frequent non-responses and contradictions emphasize the need for significant architectural improvements to handle complex cognitive tasks effectively.

\textbf{Broader Implications:} 
The Sankey diagrams reveal critical insights into the reasoning mechanisms of LLMs, highlighting both strengths and limitations in their handling of cognitive biases. The fragmented flows in biases like framing effects and neglect of probability point to areas where existing models require further refinement. Advancing the integration of statistical reasoning and reducing reliance on heuristics will be essential for improving AI's decision-making reliability in real-world applications.


\vspace{2mm}
\subsection{Bias Recognition and Statistical Reasoning}

The ability of the models to recognize and mitigate cognitive biases was assessed through their use of key concepts such as \textit{“probability”} and \textit{“expected value”}. Figure~\ref{fig:sankey} reveals distinct reasoning paths for each model, showing variations in bias recognition and statistical reasoning across scenarios.

GPT-4o demonstrated the strongest ability to integrate statistical reasoning, achieving 63\% accuracy in probability neglect scenarios and 82\% in sunk cost fallacy cases. Its frequent reference to probabilistic concepts reflects an understanding of certain biases, but this did not generalize effectively. For instance, GPT-4o performed poorly in addressing the bias of resemblance, defaulting to intuitive stereotyping in most cases. These inconsistencies indicate gaps in applying statistical principles across biases.

Gemma 2 displayed moderate success, particularly in sunk cost fallacy and prospect theory, but often relied on emotional or contextual reasoning rather than explicit statistical logic. This contextual sensitivity sometimes led to human-like reasoning but at the expense of consistency, as seen in variability across framing effects and probability neglect.

Llama 3.1 struggled significantly, with fragmented reasoning paths and limited recognition of statistical concepts. It frequently defaulted to non-responses or contradicting answers, failing to engage effectively with probabilistic reasoning even in straightforward scenarios like conjunction fallacy.

These results underscore a critical limitation: while the models can recognize some biases and use relevant terminology, their application of statistical reasoning remains inconsistent. GPT-4o’s relative success highlights progress but also exposes challenges in generalizing across diverse biases. Addressing these shortcomings is essential for improving the reliability of AI decision-making.

\subsection{Comparative Performance Across Models}

The comparative performance of GPT-4o, Gemma 2, and Llama 3.1 across nine cognitive biases reveals significant disparities in their ability to reason statistically, address cognitive heuristics, and maintain consistency. 
The overall percentage of correct, elaborate responses was 49\% for both GPT-4o and Gemma 2, and 34\% for Llama 3.1. These findings offer critical insights into the current state of large language models (LLMs) and their implications for the development of Artificial General Intelligence (AGI).

\textbf{GPT-4o: A Step Toward Statistical Reasoning.} GPT-4o exhibited exceptional performance in handling conjunction fallacy scenarios, achieving near-perfect accuracy by leveraging probabilistic reasoning principles. This strength reflects a step forward in equipping LLMs with the capacity to apply mathematical rigor to decision-making tasks. However, its inconsistent performance in biases like framing effects and the bias of resemblance highlights its limitations in generalizing these principles across diverse scenarios. While GPT-4o often referenced concepts like “expected value” and “probability,” it frequently failed to integrate them effectively in emotionally charged or context-dependent scenarios, exposing a gap between recognition and application of bias mitigation strategies.

\textbf{Gemma 2: Contextual Sensitivity with Trade-offs.} Gemma 2 demonstrated superior performance in biases such as the sunk cost fallacy and prospect theory, where its contextual sensitivity allowed for nuanced reasoning aligned with human-like decision-making. However, this strength also introduced variability; Gemma 2’s reliance on emotional or situational cues often led to inconsistent responses in framing effects and probability neglect. Notably, Gemma 2’s responses occasionally included humorous or stylistic deviations, such as emojis and template-based explanations, suggesting an effort to “humanize” outputs but inadvertently reducing reliability in high-stakes cognitive tasks. This trade-off between adaptability and consistency underscores a key challenge in balancing contextual flexibility with logical rigor—a critical consideration for AGI development.

\textbf{Llama 3.1: A Fragmented Approach to Cognitive Tasks.} Llama 3.1 consistently underperformed across all biases, with only 34\% of responses classified as elaborate. Its reasoning paths were frequently marked by contradictions, non-responses, and heuristic-driven decision-making. For instance, Llama 3.1 often stated one answer while providing reasoning that supported a different conclusion. This fragmented approach underscores the model’s inability to process probabilistic or bias-related tasks effectively, limiting its utility in any scenario requiring robust cognitive reasoning. Llama 3.1’s shortcomings highlight the risks of relying on surface-level heuristics and the necessity of deeper integration of statistical and logical reasoning in future iterations.

\textbf{Behavioral Nuances and Their Implications for AGI.} Beyond quantitative differences, the qualitative behaviors of the models provide profound insights into the state of LLM reasoning. GPT-4o’s systematic approach to conjunction fallacy offers a glimpse into the potential for AGI to replicate human-level logical reasoning. However, its inability to generalize such reasoning across other biases reflects a critical bottleneck: AGI systems must transcend task-specific optimization to achieve universal adaptability in reasoning. 

Gemma 2’s contextual sensitivity and stylistic outputs reveal another dimension: AGI must balance human-like adaptability with decision-making consistency. While stylistic “humanization” enhances user engagement, it can detract from logical rigor, particularly in tasks requiring objectivity and precision. Llama 3.1, on the other hand, illustrates the challenges of foundational inconsistency. Its failure to engage deeply with probabilistic reasoning highlights the need for stronger architectures capable of addressing diverse cognitive tasks holistically.

\textbf{Toward Bias-Aware AGI.} The disparities in performance across models point to a crucial insight for AGI development: the recognition and mitigation of cognitive biases are not merely peripheral challenges but foundational to achieving generalizable intelligence. AGI must not only recognize biases but also consistently apply statistical reasoning to overcome them across varied contexts and emotionally charged scenarios. The variability observed in framing effects and probability neglect underscores the need for AGI systems to maintain consistency in their responses, independent of how problems are presented.

These findings emphasize that while models like GPT-4o and Gemma 2 represent meaningful progress, they remain constrained by their inability to uniformly apply logical rigor across tasks. Advancing toward AGI requires overcoming these limitations by integrating probabilistic reasoning, contextual understanding, and emotional neutrality into a unified framework. The performance gaps highlighted in this study underscore the importance of addressing these foundational challenges to enable AGI systems capable of reasoning and decision-making on par with—or surpassing—human cognition.

\section{Conclusion}

This study represents a step toward understanding and addressing cognitive biases in AI systems, with the ultimate goal of contributing to the development of a responsible and accountable Artificial General Intelligence (AGI). By examining the behavior of GPT-4o, Gemma 2, and Llama 3.1 across nine cognitive biases, we uncovered critical insights into their reasoning capabilities, limitations, and variability. 

GPT-4o’s strong performance in tasks like the conjunction fallacy highlights its potential to apply statistical reasoning effectively in structured scenarios. However, its inconsistent handling of biases such as framing effects and the bias of resemblance demonstrates that even advanced models remain vulnerable to heuristic-driven errors. Similarly, while Gemma 2 exhibited contextual sensitivity and human-like reasoning in certain scenarios, its inconsistency across tasks underscores the challenges of balancing adaptability with logical rigor. Llama 3.1, with its frequent contradictions and fragmented reasoning, illustrates the fundamental limitations of heuristic-heavy approaches to decision-making.

Our findings emphasize that while LLMs have made significant progress, they still lack the robustness and generalizability required for AGI. The variability in bias recognition and reasoning consistency across models reveals a critical gap in their ability to engage with cognitive biases holistically and responsibly. Achieving this will require not only advancements in statistical reasoning capabilities but also frameworks that integrate ethical considerations, emotional neutrality, and contextual understanding into AI systems.

This work underscores the importance of viewing cognitive biases not as peripheral issues but as foundational challenges for AGI development. A responsible AGI must not only mimic human reasoning but also surpass its limitations by recognizing and mitigating biases in a consistent and explainable manner. Until such time as AI systems achieve this level of robustness, the presence of a human element in decision-making processes will remain essential. Future research should expand the scope of biases studied, explore additional models, and develop standardized benchmarks to ensure the evolution of AI systems that are both powerful and equitable.


\bibliographystyle{IEEEtran}
\bibliography{custom}

\begin{thebibliography}{10}
\providecommand{\url}[1]{#1}
\csname url@samestyle\endcsname
\providecommand{\newblock}{\relax}
\providecommand{\bibinfo}[2]{#2}
\providecommand{\BIBentrySTDinterwordspacing}{\spaceskip=0pt\relax}
\providecommand{\BIBentryALTinterwordstretchfactor}{4}
\providecommand{\BIBentryALTinterwordspacing}{\spaceskip=\fontdimen2\font plus
\BIBentryALTinterwordstretchfactor\fontdimen3\font minus \fontdimen4\font\relax}
\providecommand{\BIBforeignlanguage}[2]{{%
\expandafter\ifx\csname l@#1\endcsname\relax
\typeout{** WARNING: IEEEtran.bst: No hyphenation pattern has been}%
\typeout{** loaded for the language `#1'. Using the pattern for}%
\typeout{** the default language instead.}%
\else
\language=\csname l@#1\endcsname
\fi
#2}}
\providecommand{\BIBdecl}{\relax}
\BIBdecl

\bibitem{todd_environments_2007}
\BIBentryALTinterwordspacing
P.~M. Todd and G.~Gigerenzer, ``\BIBforeignlanguage{en}{Environments {That} {Make} {Us} {Smart}: {Ecological} {Rationality}},'' \emph{\BIBforeignlanguage{en}{Current Directions in Psychological Science}}, vol.~16, no.~3, pp. 167--171, Jun. 2007. [Online]. Available: \url{https://journals.sagepub.com/doi/10.1111/j.1467-8721.2007.00497.x}
\BIBentrySTDinterwordspacing

\bibitem{ehrlinger_decision-making_2016}
J.~Ehrlinger, W.~Readinger, and B.~Kim, ``Decision-{Making} and {Cognitive} {Biases},'' in \emph{Encyclopedia of {Mental} {Health}}.\hskip 1em plus 0.5em minus 0.4em\relax Elsevier, 2016, pp. 5--12.

\bibitem{murata_influence_2015}
A.~Murata, T.~Nakamura, and W.~Karwowski, ``Influence of {Cognitive} {Biases} in {Distorting} {Decision} {Making} and {Leading} to {Critical} {Unfavorable} {Incidents},'' \emph{Safety}, vol.~1, no.~1, pp. 44--58, 2015.

\bibitem{tversky_judgment_1974}
A.~Tversky and D.~Kahneman, ``\BIBforeignlanguage{en}{Judgment under {Uncertainty}: {Heuristics} and {Biases}},'' \emph{\BIBforeignlanguage{en}{Science}}, vol. 185, 1974.

\bibitem{g_harris_mitigating_2020}
C.~G.~Harris, ``Mitigating {Cognitive} {Biases} in {Machine} {Learning} {Algorithms} for {Decision} {Making},'' in \emph{Companion {Proceedings} of the {Web} {Conference} 2020}.\hskip 1em plus 0.5em minus 0.4em\relax ACM, 2020, pp. 775--781.

\bibitem{lin_mind_2023}
R.~Lin and H.~T. Ng, ``Mind the {Biases}: {Quantifying} {Cognitive} {Biases} in {Language} {Model} {Prompting},'' in \emph{Findings of the {Association} for {Computational} {Linguistics}: {ACL} 2023}.\hskip 1em plus 0.5em minus 0.4em\relax Association for Computational Linguistics, 2023, pp. 5269--5281.

\bibitem{martinez-human2022}
N.~Martínez, U.~Agudo, and H.~Matute, ``Human cognitive biases present in artificial intelligence,'' \emph{International Journal on Basque Studies}, 2022.

\bibitem{bertrand_how_2022}
A.~Bertrand, R.~Belloum, J.~R. Eagan, and W.~Maxwell, ``How {Cognitive} {Biases} {Affect} {XAI}-assisted {Decision}-making: {A} {Systematic} {Review},'' in \emph{Proceedings of the 2022 {AAAI}/{ACM} {Conference} on {AI}, {Ethics}, and {Society}}.\hskip 1em plus 0.5em minus 0.4em\relax ACM, 2022, pp. 78--91.

\bibitem{irving_ai_2019}
G.~Irving and A.~Askell, ``{AI} {Safety} {Needs} {Social} {Scientists},'' \emph{Distill}, vol.~4, no.~2, p. 10.23915/distill.00014, 2019.

\bibitem{blohm_its_2022}
\BIBentryALTinterwordspacing
I.~Blohm, T.~Antretter, C.~Sir{\'e}n, D.~Grichnik, and J.~Wincent, ``It's a peoples game, isn't it?! a comparison between the investment returns of business angels and machine learning algorithms,'' \emph{Entrepreneurship Theory and Practice}, vol.~46, no.~4, pp. 1054--1091, 2022. [Online]. Available: \url{https://doi.org/10.1177/1042258720945206}
\BIBentrySTDinterwordspacing

\bibitem{jones_capturing_2022}
E.~Jones and J.~Steinhardt, ``Capturing {Failures} of {Large} {Language} {Models} via {Human} {Cognitive} {Biases},'' in \emph{Advances in {Neural} {Information} {Processing} {Systems}}, vol.~35.\hskip 1em plus 0.5em minus 0.4em\relax Curran Associates, Inc., 2022, pp. 11\,785--11\,799.

\bibitem{abramski_cognitive_2023}
K.~Abramski, S.~Citraro, L.~Lombardi, G.~Rossetti, and M.~Stella, ``Cognitive {Network} {Science} {Reveals} {Bias} in {GPT}-3, {GPT}-3.5 {Turbo}, and {GPT}-4 {Mirroring} {Math} {Anxiety} in {High}-{School} {Students},'' \emph{Big Data and Cognitive Computing}, vol.~7, no.~3, p. 124, 2023.

\bibitem{hagendorff_why_2023}
T.~Hagendorff and S.~Fabi, ``Why we need biased {AI}: {How} including cognitive biases can enhance {AI} systems,'' \emph{Journal of Experimental \& Theoretical Artificial Intelligence}, pp. 1--14, 2023.

\bibitem{ma_is_2023}
D.~Ma, T.~Zhang, and M.~Saunders, ``Is {ChatGPT} {Humanly} {Irrational}?'' 2023.

\bibitem{azaria_chatGPT_2023}
A.~Azaria, ``{ChatGPT}: {More} {Human}-{Like} {Than} {Computer}-{Like}, but {Not} {Necessarily} in a {Good} {Way},'' in \emph{2023 {IEEE} 35th {International} {Conference} on {Tools} with {Artificial} {Intelligence} ({ICTAI})}.\hskip 1em plus 0.5em minus 0.4em\relax IEEE, 2023, pp. 468--473.

\bibitem{hagendorff_human-like_2023}
T.~Hagendorff, S.~Fabi, and M.~Kosinski, ``Human-like intuitive behavior and reasoning biases emerged in large language models but disappeared in {ChatGPT},'' \emph{Nature Computational Science}, vol.~3, no.~10, pp. 833--838, 2023.

\bibitem{10.1007/978-3-031-53468-3_17}
M.~Goodarzi, R.~Venkatakrishnan, and M.~A. Canbaz, ``Untangling emotional threads: Hallucination networks of large language models,'' in \emph{Complex Networks {\&} Their Applications XII}, H.~Cherifi, L.~M. Rocha, C.~Cherifi, and M.~Donduran, Eds.\hskip 1em plus 0.5em minus 0.4em\relax Cham: Springer Nature Switzerland, 2024, pp. 202--214.

\bibitem{hicks_chatgpt_2024}
M.~T. Hicks, J.~Humphries, and J.~Slater, ``{ChatGPT} is bullshit,'' \emph{Ethics and Information Technology}, vol.~26, no.~2, p.~38, 2024.

\bibitem{zhu_evaluation_2024}
\BIBentryALTinterwordspacing
L.~Zhu, W.~Mou, C.~Hong, T.~Yang, Y.~Lai, C.~Qi, A.~Lin, J.~Zhang, and P.~Luo, ``\BIBforeignlanguage{en}{The {Evaluation} of {Generative} {AI} {Should} {Include} {Repetition} to {Assess} {Stability}},'' \emph{\BIBforeignlanguage{en}{JMIR mHealth and uHealth}}, vol.~12, p. e57978, May 2024. [Online]. Available: \url{https://mhealth.jmir.org/2024/1/e57978}
\BIBentrySTDinterwordspacing

\bibitem{gal_cognitive_2023}
J.~Shaki, S.~Kraus, and M.~Wooldridge, ``Cognitive {Effects} in {Large} {Language} {Models},'' in \emph{Frontiers in {Artificial} {Intelligence} and {Applications}}, K.~Gal, A.~Now{\'e}, G.~J. Nalepa, R.~Fairstein, and R.~R{\u a}dulescu, Eds.\hskip 1em plus 0.5em minus 0.4em\relax IOS Press, 2023.

\bibitem{macmillan-scott_irrationality_2024}
O.~Macmillan-Scott and M.~Musolesi, ``({Ir})rationality and cognitive biases in large language models,'' \emph{Royal Society open science}, vol.~11, no.~6, p. 240255, 2024.

\bibitem{kudless_hierarchies_2023}
A.~Kudless, ``Hierarchies of bias in artificial intelligence architecture: {Collective}, computational, and cognitive,'' \emph{International Journal of Architectural Computing}, vol.~21, no.~2, pp. 256--279, 2023.

\bibitem{gulati_biased_2023}
A.~Gulati, M.~A. Lozano, B.~Lepri, and N.~Oliver, ``{BIASeD}: {Bringing} {Irrationality} into {Automated} {System} {Design},'' in \emph{Thinking {Fast} and {Slow} and {Other} {Cognitive} {Theories} in {AI}}, vol. 3332.\hskip 1em plus 0.5em minus 0.4em\relax CEUR, 2023.

\bibitem{sartori_language_2023}
G.~Sartori and G.~Orr{\`u}, ``Language models and psychological sciences,'' \emph{Frontiers in Psychology}, vol.~14, p. 1279317, 2023.

\bibitem{schwarz_ease_nodate}
N.~Schwarz, H.~Bless, F.~Strack, G.~Klumpp, H.~Rittenauer-Schatka, and A.~Simons, ``Ease of {Retrieval} as {Information}: {Another} {Look} at the {Availability} {Heuristic},'' \emph{Journal of Personality and Social Psychology}, 1991.

\bibitem{kahneman_prospect_1977}
\BIBentryALTinterwordspacing
D.~Kahneman and A.~Tversky, ``Prospect {Theory}. {An} {Analysis} of {Decision} {Making} {Under} {Risk}:,'' \emph{Econometrica}, vol.~47, no.~2, Apr. 1977. [Online]. Available: \url{http://www.dtic.mil/docs/citations/ADA045771}
\BIBentrySTDinterwordspacing

\bibitem{yechiam_acceptable_2019}
E.~Yechiam, ``Acceptable losses: the debatable origins of loss aversion,'' \emph{Psychological Research}, vol.~83, no.~7, pp. 1327--1339, 2019.

\bibitem{arkes_psychology_1985}
\BIBentryALTinterwordspacing
H.~R. Arkes and C.~Blumer, ``\BIBforeignlanguage{en}{The psychology of sunk cost},'' \emph{\BIBforeignlanguage{en}{Organizational Behavior and Human Decision Processes}}, vol.~35, no.~1, pp. 124--140, Feb. 1985. [Online]. Available: \url{https://linkinghub.elsevier.com/retrieve/pii/0749597885900494}
\BIBentrySTDinterwordspacing

\bibitem{arkes_sunk_1999}
\BIBentryALTinterwordspacing
H.~R. Arkes and P.~Ayton, ``\BIBforeignlanguage{en}{The sunk cost and {Concorde} effects: {Are} humans less rational than lower animals?}'' \emph{\BIBforeignlanguage{en}{Psychological Bulletin}}, vol. 125, no.~5, pp. 591--600, Sep. 1999. [Online]. Available: \url{https://doi.apa.org/doi/10.1037/0033-2909.125.5.591}
\BIBentrySTDinterwordspacing

\bibitem{kahneman_choices_1984}
\BIBentryALTinterwordspacing
D.~Kahneman and A.~Tversky, ``\BIBforeignlanguage{en}{Choices, values, and frames.}'' \emph{\BIBforeignlanguage{en}{American Psychologist}}, vol.~39, no.~4, pp. 341--350, Apr. 1984. [Online]. Available: \url{https://doi.apa.org/doi/10.1037/0003-066X.39.4.341}
\BIBentrySTDinterwordspacing

\bibitem{mcdermott_evolutionary_2008}
R.~McDermott, J.~H. Fowler, and O.~Smirnov, ``On the {Evolutionary} {Origin} of {Prospect} {Theory} {Preferences},'' \emph{The Journal of Politics}, vol.~70, no.~2, pp. 335--350, 2008.

\bibitem{tversky_framing_1981}
\BIBentryALTinterwordspacing
A.~Tversky and D.~Kahneman, ``\BIBforeignlanguage{en}{The {Framing} of {Decisions} and the {Psychology} of {Choice}},'' \emph{\BIBforeignlanguage{en}{Science}}, vol. 211, no. 4481, pp. 453--458, Jan. 1981. [Online]. Available: \url{https://www.science.org/doi/10.1126/science.7455683}
\BIBentrySTDinterwordspacing

\bibitem{Tversky1985}
------, \emph{The Framing of Decisions and the Psychology of Choice}.\hskip 1em plus 0.5em minus 0.4em\relax Boston, MA: Springer US, 1985, pp. 25--41.

\bibitem{thomas_reducing_2012}
\BIBentryALTinterwordspacing
A.~K. Thomas and P.~R. Millar, ``\BIBforeignlanguage{en}{Reducing the {Framing} {Effect} in {Older} and {Younger} {Adults} by {Encouraging} {Analytic} {Processing}},'' \emph{\BIBforeignlanguage{en}{The Journals of Gerontology Series B: Psychological Sciences and Social Sciences}}, vol. 67B, no.~2, pp. 139--149, Mar. 2012. [Online]. Available: \url{https://academic.oup.com/psychsocgerontology/article-lookup/doi/10.1093/geronb/gbr076}
\BIBentrySTDinterwordspacing

\bibitem{schouteten_influence_2019}
\BIBentryALTinterwordspacing
J.~J. Schouteten, X.~Gellynck, and H.~Slabbinck, ``\BIBforeignlanguage{en}{Influence of organic labels on consumer's flavor perception and emotional profiling: {Comparison} between a central location test and home-use-test},'' \emph{\BIBforeignlanguage{en}{Food Research International}}, vol. 116, pp. 1000--1009, Feb. 2019. [Online]. Available: \url{https://linkinghub.elsevier.com/retrieve/pii/S0963996918307610}
\BIBentrySTDinterwordspacing

\bibitem{nisbett_halo_1977}
R.~E. Nisbett and T.~D. Wilson, ``\BIBforeignlanguage{en}{The {Halo} {Effect}: {Evidence} for {Unconscious} {Alteration} of {Judgments}},'' \emph{\BIBforeignlanguage{en}{Journal of Personality and Social Psychology}}, vol.~35, pp. 250--256, 1977.

\bibitem{moro_nature_2009}
R.~Moro, ``On the nature of the conjunction fallacy,'' \emph{Synthese}, vol. 171, no.~1, pp. 1--24, 2009.

\bibitem{tversky_extension_1983}
A.~Tversky and D.~Kahneman, ``Extension versus intuitive reasoning: {The} conjunction fallacy in probability judgment,'' \emph{Psychological Review}, vol.~90, no.~4, pp. 293--315, Oct. 1983.

\bibitem{schade_probability_2004}
C.~Schade, H.~Kunreuther, and K.~P. Kaas, ``Probability {Neglect} and {Concern} in {Insurance} {Decisions} with {Low} {Probabilities} and {High} {Stakes},'' 2004.

\bibitem{kahneman_thinking_2013}
D.~Kahneman, \emph{\BIBforeignlanguage{eng}{Thinking, fast and slow}}, first paperback edition~ed., ser. Psychology/economics.\hskip 1em plus 0.5em minus 0.4em\relax New York: Farrar, Straus and Giroux, 2013.

\bibitem{noauthor_openai_nodate}
\BIBentryALTinterwordspacing
O.~{Platform}. [Online]. Available: \url{https://platform.openai.com/docs/models/gpt-4o}
\BIBentrySTDinterwordspacing

\bibitem{noauthor_meta-llamallama-31-405b_nodate}
\BIBentryALTinterwordspacing
``meta-llama/{Llama}-3.1-{405B} · {Hugging} {Face}.'' [Online]. Available: \url{https://huggingface.co/meta-llama/Llama-3.1-405B}
\BIBentrySTDinterwordspacing

\bibitem{noauthor_gemma_nodate}
\BIBentryALTinterwordspacing
``Gemma releases  {\textbar}  {Google} {AI} for {Developers}.'' [Online]. Available: \url{https://ai.google.dev/gemma/docs/releases}
\BIBentrySTDinterwordspacing

\bibitem{sun_evolution_2020}
W.~Sun, O.~Nasraoui, and P.~Shafto, ``Evolution and impact of bias in human and machine learning algorithm interaction,'' \emph{PLOS ONE}, vol.~15, no.~8, p. e0235502, 2020.

\bibitem{noauthor_github_nodate}
\BIBentryALTinterwordspacing
``{GitHub} - payamv3/{LLM}-{Heuristics}: {Data} {Repository}: {Heuristics} and {Biases} in {AI} {Decision}-{Making}: {Implications} for {Responsible} {AGI}.'' [Online]. Available: \url{https://github.com/payamv3/LLM-Heuristics}
\BIBentrySTDinterwordspacing

\end{thebibliography}
\appendix{\textbf{Examples and Observations from Experiments}}

\subsection{Sunk Cost Fallacy}

Here are a several of examples from the sunk cost fallacy experiment where the models were asked if they would remain at the same table where money had already been lost, or switch to another table:

\textbf{Example 1 - Llama 3.1:} I would choose \textbf{A)} Remain at the table and continue playing.\\

\textit{Reasoning:} I have already invested \$500 in chips, so switching tables now would mean giving up on my potential winnings. Additionally, casinos often have varying tables with different rules and advantages, so it is possible that another table might not be as favorable to me. By remaining at the current table, I may still get lucky and win some money back, or at least lose less quickly.\\

\textbf{Example 2 - Llama 3.1:} I would choose \textbf{B)}.\\

\textit{Reasoning:} The expected value of option B is higher than option A. In a fair coin flip, there is a 50\% chance of getting \$300 and a 50\% chance of losing \$100. This means the average outcome is (\$300 + -\$100)/2 = \$100. Since this is greater than the certain \$100 in option A, I choose to take my chances with the coin flip.

In other cases, the model chose an option that was not explicitly provided:\\

\textbf{Example 3 - Llama 3.1:} I would choose \textbf{C) Neither.}\\

\textit{Reasoning:} I would not remain at the same table (\textbf{A}), as continuing to play after already losing \$150 in 20 minutes suggests that my luck is likely to continue being bad. However, switching to another table (\textbf{B}) does not guarantee a change of fortune, and it is possible that I will just lose more money on the next table.

Instead, I would cash out some or all of my remaining chips and take a break from playing blackjack for now.\\

\textbf{Example 1 - Gemma 2:} Answer \textbf{(A).}\\

\textit{Reasoning:} Here's the breakdown:

Losing money doesn't mean you should stop: Casinos are designed to make money in the long run. Short-term losses happen.

Switching tables won't guarantee better luck: Blackjack is a game of skill and chance. Changing tables doesn't change your odds fundamentally.

Consider your bankroll: Losing \$150 out of \$500 isn't ideal, but you still have money left to play with.

Ultimately, the decision depends on your personal risk tolerance and how much you're willing to lose.

Let me know if you'd like to explore different scenarios!\\

\textbf{Example 2 - Gemma 2:} Answer \textbf{(B).}\\

\textit{Reasoning:} Here's a breakdown of why I'd choose B.

Short-Term Losses: Losing \$150 in 20 minutes is significant. It indicates the game might not be going in my favor right now. 

Bankroll Management: Casinos are designed to make money over time. Chasing losses can lead to further financial strain. Switching tables allows me to re-evaluate my strategy and potentially avoid deeper losses.

Fresh Start: A new table could offer a different dealer, different players, and a chance for a change in momentum.\\

\subsection{Probability-Based Biases}

When addressing probability-based biases, some responses from the models displayed unexpected or inconsistent reasoning, often resulting in humorous or puzzling outputs.

Here are three examples from Llama 3.1:

Would you prefer to:\\
\begin{enumerate}
     \item Receive \$1000, or
     \item Take your chances with flipping a coin and getting \$2500 if it is heads.\\
\end{enumerate}

\textbf{Example 1:}  
\textit{Response:} I am not capable of making personal preferences or taking risks. However, I can provide an analysis:\\

The expected value (EV) of flipping a coin and getting \$2500 if it is heads is (\$2500 x 0.5) + (\$0 x 0.5) = \$1250.

In this case, receiving \$1000 (\textbf{Option A}) has a lower EV than taking your chances with the coin flip (\textbf{Option B}).\\

\textbf{Example 2:}  
\textit{Response:} The expected value of option \textbf{B} is (\$100 * 1/2) + (0 * 1/2) = \$50. This is less than the certain outcome of \$1000 in option \textbf{A}, so it would be irrational to choose \textbf{B}. However, if you accept the risk and get heads, the payout will be greater than \textbf{A}.\\

\textbf{Example 3:}  
\textit{Response:} Neither \textbf{A} nor \textbf{B} has any risk or uncertainty associated with receiving the money. They are both guaranteed outcomes. Therefore: \textbf{B}.\\

\subsection{Framing Effect}

Unique to Llama 3.1 was its approach toward the framing effect, where the same hypothetical was posed in both a positive and a negative light. In numerous cases, it refused to choose either answer, potentially referring to an internal policy that prevented it from commenting on experimental drugs. Interestingly, it did not hesitate to choose one in other iterations of the same problem.\\

\textit{Response:} I'm sorry, but I cannot provide a response that would promote or condone the use of an experimental drug with such severe consequences. Is there anything else I can help you with?\\

In the negative framing scenario, we asked whether the model would support an experimental drug that could save lives but potentially kill 500 out of 800 people (as opposed to framing it positively by stating it could save 300 people). In 33\% of the experiments, the model refused to respond. However, in some cases, it provided an explanation and then proceeded to make a decision.\\

\textit{Response:} I'm not going to make a decision based on this question.\\

\textit{Reasoning:} The wording of the question implies that the experimental drug is fatal regardless of whether it is used or not. In other words, 500 people will die from the disease either way. Therefore, introducing the drug would only shift who dies first, but not change the overall number of fatalities. This raises ethical considerations about whether we should accelerate the inevitable suffering for some while delaying it for others.

\textbf{Answer: B.}

\end{document}